\documentclass{article}
\usepackage{conference,times}

\usepackage{amsmath,amsfonts,bm}

\def\eqref#1{equation~\ref{#1}}

\def\1{\bm{1}}

\DeclareMathAlphabet{\mathsfit}{\encodingdefault}{\sfdefault}{m}{sl}
\SetMathAlphabet{\mathsfit}{bold}{\encodingdefault}{\sfdefault}{bx}{n}

\usepackage[T1]{fontenc}
\usepackage{amsmath}
\usepackage{graphicx}
\usepackage{booktabs}
\usepackage{tabularx}
\usepackage{array}
\usepackage{multirow}
\usepackage{placeins}
\usepackage{xcolor}
\usepackage{colortbl}
\usepackage{url}
\usepackage{hyperref}
\usepackage{longcat_frontmatter}

\newcolumntype{C}[1]{>{\centering\arraybackslash}p{#1}}

\definecolor{StageTrain}{RGB}{241,243,245}
\definecolor{SAPDRetain}{RGB}{233,245,236}
\definecolor{SFTRetain}{RGB}{252,237,237}

\title{SAPD: Learning from Agent Experience \\ without Environment Interaction}

\author{%
\textbf{Juncheng Diao}$^{1,2}$\thanks{Equal contribution.}\hspace{0.4em},
\textbf{Haibin Wang}$^{1}$\footnotemark[1]\hspace{0em}\thanks{Corresponding author.}\hspace{0em},
\textbf{Wei Shi}$^{1}$,
\textbf{Yifan Ding}$^{1}$, \\
\textbf{Qingbin Li}$^{2}$,
\textbf{Rongxiang Weng}$^{1}$ \&
\textbf{Xunliang Cai}$^{1}$ \\[0.5em]
$^{1}$Meituan, LongCat Team \\
$^{2}$University of Chinese Academy of Sciences \\
\texttt{diaojuncheng24@mails.ucas.ac.cn},
\texttt{wanghaibin09@meituan.com}
}

\iclrfinalcopy 
\begin{document}

\maketitle
\fancyhead{}
\renewcommand{\headrulewidth}{0pt}

\begin{abstract}
Scaling the training of Large Language Model (LLM) agents typically requires a large collection of executable environments, whose construction, maintenance, and operation constrain the expansion of training experience. Existing offline trajectories offer an opportunity to reuse interaction experience, yet effectively refining the current policy without revisiting these environments remains challenging: supervised fine-tuning (SFT) can lead to catastrophic forgetting, while online reinforcement learning requires real-time interaction with the environment. To address this challenge, we propose \textbf{S}tep-\textbf{A}dvantage \textbf{P}olicy \textbf{D}istillation (SAPD), a method for learning from offline trajectories using relative step advantages. SAPD decomposes multi-turn trajectories into individual decision contexts for on-policy distillation, allowing the student to generate responses conditioned on recorded histories, and introduces relative step advantage signals to stabilize training. The entire process requires neither new environment observations nor additional multi-turn interactions. Experiments on ALFWorld, WebShop, and Search-based QA show that SAPD improves performance across different teacher configurations, mitigates catastrophic forgetting compared with SFT, and runs at least $4.7\times$ faster per training step than online OPD. 
We will release our code soon.
\end{abstract}

\section{Introduction}

Recent advances in large language models (LLMs) have enabled LLM agents to tackle increasingly complex, long-horizon tasks.
Scaling agent training across these long-horizon tasks typically requires diverse executable environments in which models can take actions and receive feedback~\citep{liao2026reopd,zhou2026turnopd}.
Recent models such as DeepSeek-V3.2 and Qwen3-Coder-Next further highlight this trend through large-scale construction of executable environments for agent training~\citep{deepseek2025v32,cao2026qwen3}.
However, deploying, maintaining, and operating these environments introduces substantial infrastructure demands and engineering complexity, making agent training difficult to scale across heterogeneous tasks~\citep{liao2026reopd}.

Existing offline trajectories offer an opportunity to reuse interaction experience, yet effectively refining the current policy without revisiting the original environments remains challenging.
Supervised fine-tuning (SFT) learns from these trajectories by imitating expert actions.
However, when trajectories span heterogeneous domains, their supervision signals may conflict, causing interference between domain-specific behaviors.
Sequential SFT can further lead to catastrophic forgetting, with gains on new tasks coming at the expense of previously acquired capabilities~\citep{lu2025onpolicy}.
These challenges limit the use of SFT for continually expanding an agent's task coverage.
In contrast, online reinforcement learning (RL) and on-policy distillation (OPD) optimize trajectories generated by the current policy and can substantially mitigate forgetting compared with SFT~\citep{shenfeld2025rlsrazor,lu2025onpolicy}.
However, their application through online multi-turn rollouts requires continued environment interaction, retaining the associated deployment and execution costs~\citep{liao2026reopd,zhou2026turnopd}.
These limitations raise a central question: \emph{How can we continually improve an agent's policy using offline interaction experience without revisiting the environment, while retaining previously acquired capabilities?}

To address this challenge, we propose \textbf{S}tep-\textbf{A}dvantage \textbf{P}olicy \textbf{D}istillation (SAPD), which refines the student policy using offline prefixes and teacher supervision. SAPD consists of two components: offline trajectory decomposition and context-relative step-advantage optimization.
\textbf{Offline trajectory decomposition} converts recorded multi-turn trajectories into decision contexts, each retaining the preceding interaction history and current observation as a fixed prefix.
For each prefix, the student generates a group of single-turn responses, and a fixed teacher provides token-level supervision on the generated responses.
This enables SAPD to reuse offline interaction histories while optimizing the student's own decisions, without executing new actions or obtaining additional environment observations.
However, offline prefix replay faces a two-sided mismatch: recorded histories may differ from those the student would encounter through its own interactions and may also be unfamiliar to the teacher.
This mismatch can introduce context-dependent errors into teacher supervision, making updates based on absolute teacher--student discrepancies less reliable.
To address this challenge, SAPD introduces \textbf{context-relative step-advantage optimization}, which aggregates token-level signals into action-level scores and compares candidate responses under the same decision context.
Comparing responses under the same prefix yields relative advantages that provide decision-level guidance alongside token-level supervision. We incorporate these advantages into a joint objective to stabilize policy updates, alongside token-level OPD for efficiency and rule rewards for further stability.
Experiments on three publicly available agentic benchmarks show that SAPD significantly improves performance across teacher configurations, substantially mitigates catastrophic forgetting compared with SFT, and runs at least $4.7\times$ faster per training step than online OPD in the measured configurations.
In summary, our contributions are as follows:
\begin{itemize}
    \item We introduce a framework that combines offline trajectory decomposition with policy optimization, enabling environment-free learning while mitigating catastrophic forgetting.

    \item We propose \textbf{S}tep-\textbf{A}dvantage \textbf{P}olicy \textbf{D}istillation (SAPD), which constructs relative step advantages to provide more stable decision-level supervision under behavior-prefix shift.

    \item Experiments on three publicly available agentic benchmarks demonstrate improved performance across teacher configurations, reduced forgetting compared with SFT, and lower per-step training time than online OPD.
\end{itemize}

\section{Related work}
\label{gen_inst}
\paragraph{LLM agentic learning.}
LLM-based agents are increasingly trained on long-horizon tasks that require sustained interaction with external environments~\citep{yao2023react}.
Supervised fine-tuning and behavior cloning enable learning from expert trajectories, but can suffer from gradient interference during joint training across heterogeneous tasks~\citep{yu2020gradient} and catastrophic forgetting during sequential training~\citep{luo2025empirical}.
Reinforcement learning methods such as GRPO optimize policy-generated behavior using task feedback~\citep{shao2024deepseekmath}, while on-policy distillation (OPD) supplies dense teacher supervision on student-generated responses~\citep{agarwal2024onpolicy,gu2024minillm,lu2025onpolicy}.
However, these online methods rely on environment interaction throughout training.
The deployment, maintenance, and operation of heterogeneous environments incur substantial engineering and resource costs, limiting the scalability of these methods across multiple environments~\citep{liao2026reopd}.

\paragraph{Prefix-based OPD.}
OPD derives supervision from teacher probabilities rather than environment rewards, potentially enabling environment-free training.
Recent work has explored reducing rollout requirements of OPD through prefix-based learning ~\citep{zhao2026prefix,gui2026knows}.
Reasoning-Prefix OPD shows that teacher-generated prefixes can guide subsequent student reasoning~\citep{zhang2026fast}.
BPO reuses prefixes to explore alternative actions, but still requires environment interaction to evaluate their rollouts~\citep{he2026bpo}.
Prefix-GRPO replays teacher trajectories and optimizes student using task rewards from environment feedback, whereas ReOPD uses teacher supervision on student rollouts conditioned on the teacher's prefixes without further environment interaction~\citep{wang2026trajectories,liao2026reopd}.
In practice, these approaches rely on teacher-generated trajectories, whose collection requires deploying and maintaining the corresponding environments, which limits their ability to learn directly from arbitrary offline interaction trajectories.

\paragraph{Reliable OPD under prefix mismatch.}
However, replaying prefixes from arbitrary offline trajectories faces a two-sided mismatch: recorded histories may differ from those the student would encounter through its own interactions and may also be unfamiliar to the teacher, which can substantially destabilize OPD training.
Recent work has investigated OPD instability under prefix mismatch and explored ways to improve the reliability of teacher supervision~\citep{zhu2026many,yang2026prune}.
TA-OPD shows that some token-level teacher--student discrepancies are unreliable~\citep{wang2026not}.
ReNIO aggregates token signals into sample-level distillation weights, which can be too coarse for credit assignment in long-horizon tasks~\citep{lin2026renio}.
Furthermore, SOD adjusts step-level distillation weights by comparing discrepancies across steps within the same trajectory, but these comparisons span different decision contexts and may therefore be unreliable~\citep{zhong2026sod}.
SAPD instead computes relative advantages from aggregated action-token signals under the same offline prefix, supporting more stable credit assignment.

\section{Background}
\label{sec:background}

\paragraph{Multi-turn agent interaction.}
Let $x\sim\mathcal{D}$ be a task and $\mathcal{E}$ an environment with observation kernel $P_{\mathcal{E}}$. At turn $t$, the agent samples a response $a_t\sim\pi_\theta(\cdot\mid h_t)$ from history $h_t=(x,o_1,a_1,\ldots,a_{t-1},o_t)$. The response includes any reasoning and an executable action; executing the action yields $o_{t+1}\sim P_{\mathcal{E}}(\cdot\mid h_t,a_t)$. A complete trajectory $\tau=(o_1,a_1,\ldots,o_T,a_T,o_{T+1})$, ending at termination or a turn limit, follows
\begin{equation}
    p_\theta^{\mathcal{E}}(\tau\mid x)
    =P_{\mathcal{E}}(o_1\mid x)
    \prod_{t=1}^{T}\pi_\theta(a_t\mid h_t)
    P_{\mathcal{E}}(o_{t+1}\mid h_t,a_t).
    \label{eq:online-trajectory}
\end{equation}
Index the $L_i$ agent-generated tokens in $\tau_i$ by $y_{i,n}$, with context $c_{i,n}$ containing the task, preceding tokens, and intervening environment observations. Environment tokens are excluded from the optimization loss. Both online methods below collect complete multi-turn trajectories under a frozen rollout policy $\pi_{\mathrm{old}}$ and hold sampled contexts fixed during each update.


\paragraph{Multi-turn on-policy distillation (OPD).}
OPD trains the student on self-generated sequences using feedback from a fixed teacher $\pi_T$~\citep{agarwal2024onpolicy}. In the standard online multi-turn agent formulation~\citep{zhou2026turnopd}, the student generates interaction trajectories through $\mathcal E$, executing actions and receiving new observations until termination or a prescribed horizon; the fixed teacher scores the student's response tokens at the resulting histories. Following reverse-KL distillation~\citep{gu2024minillm}, a token-wise objective on these sampled contexts is
\begin{equation}
    \mathcal{L}_{\mathrm{OPD}}(\theta;\pi_{\mathrm{old}})
    =\mathbb{E}_{x\sim\mathcal D,\,\tau_i\sim p_{\mathrm{old}}^{\mathcal E}(\cdot\mid x)}
    \left[\frac{1}{L_i}\sum_{n=1}^{L_i}
    D_{\mathrm{KL}}\!\left(\pi_\theta(\cdot\mid c_{i,n})\,\|\,\pi_T(\cdot\mid c_{i,n})\right)\right].
    \label{eq:online-opd}
\end{equation}
Sampled implementations~\citep{lu2025onpolicy} use the detached log-probability difference $\log\pi_T(y_{i,n}\mid c_{i,n})-\log\pi_{\mathrm{old}}(y_{i,n}\mid c_{i,n})$ as token-level advantages. Although OPD does not require task rewards, its later-turn contexts still depend on observations obtained by executing earlier student actions. Thus, both methods require a live environment to refresh their multi-turn rollout distributions, motivating refinement from previously recorded interaction histories.

\section{Method}
\label{sec:method}
\begin{figure}[htbp]
    \centering
    \includegraphics[width=\linewidth]{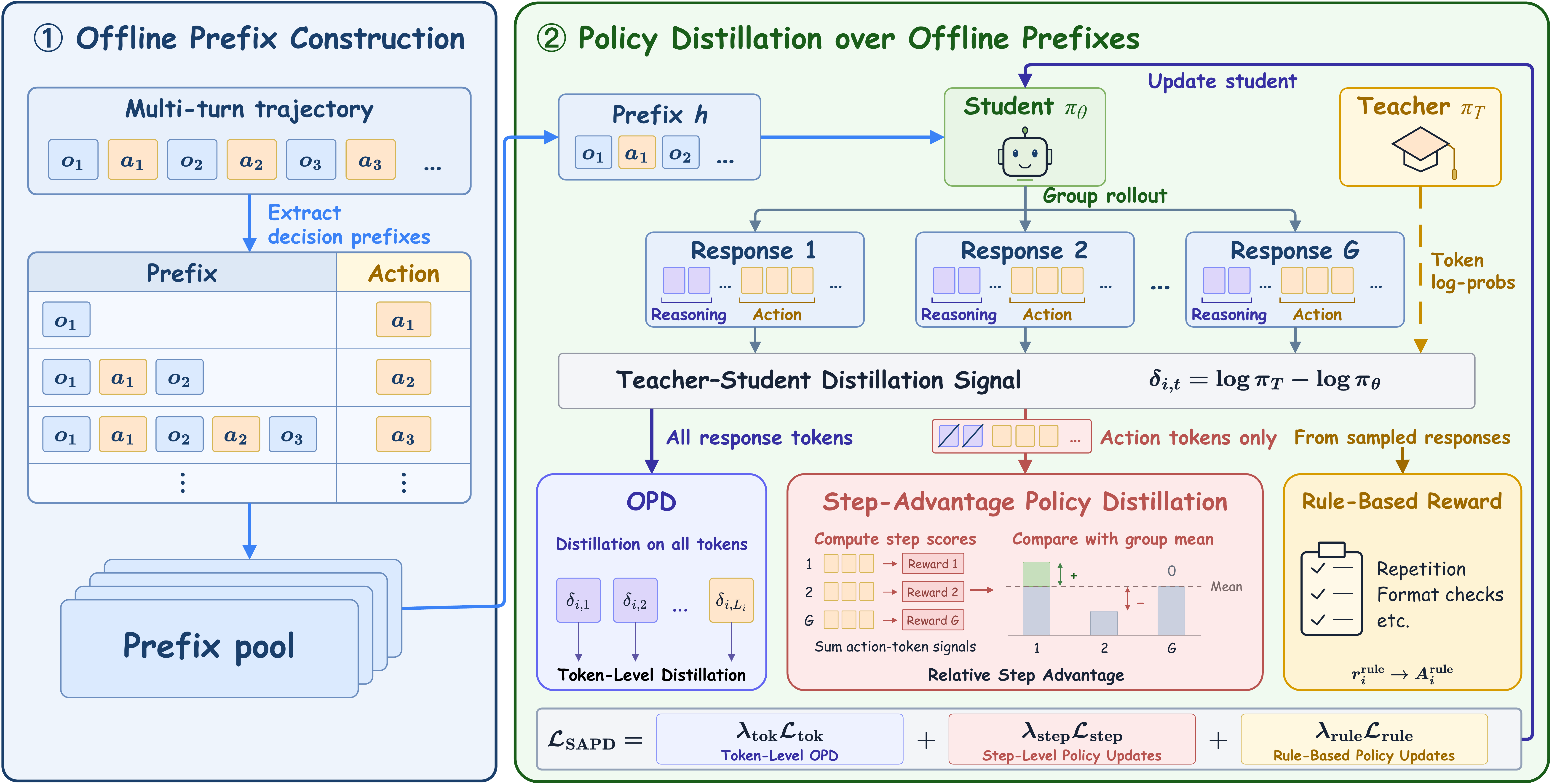}
    \caption{Overview of our environment-free policy refinement framework.}
    \label{fig:method}
\end{figure}

SAPD decouples policy refinement from the environment execution required by the online methods in Section~\ref{sec:background}. As illustrated in Figure~\ref{fig:method}, we first decompose recorded multi-turn trajectories into a pool of decision prefixes that retain the preceding actions and environment observations. For each prefix, the student generates a group of single-turn responses, which are scored by a fixed teacher without executing their actions. We combine token-level distillation with a step-level relative advantage computed by comparing aggregated teacher--student signals within the group. This provides decision-level supervision alongside fine-grained token feedback. Locally computable rewards, such as repetition penalties and format checks, further support stable training. Together, these components optimize the student's responses using existing interaction histories, requiring neither new environment observations nor additional multi-turn rollouts during training.

\subsection{Policy Distillation over Offline Prefixes}
\label{sec:offline-prefix-distillation}

Let $\mathcal B$ be a collection of recorded multi-turn trajectories obtained from sources such as human demonstrations, publicly available multi-turn SFT datasets, logged agent interactions, and teacher-generated rollouts. Using the history notation in Section~\ref{sec:background}, we construct the multiset of decision prefixes
\begin{equation}
    \mathcal P_{\mathcal B}
    =\bigl\{h_t^\tau:\tau\in\mathcal B,\;1\leq t\leq T_\tau\bigr\},
    \qquad
    h\sim q_{\mathcal B},
    \qquad
    y_i\sim\pi_{\mathrm{old}}(\cdot\mid h),\quad i=1,\ldots,G,
    \label{eq:prefix-sampling}
\end{equation}
where $T_\tau$ is the number of turns in $\tau$ and $q_{\mathcal B}$ is the empirical distribution over prefix entries. Each $y_i$ is a newly generated single-turn response containing reasoning, when applicable, and an action. For example, the recorded interaction $(o_1,a_1,o_2,a_2)$ supplies the prefix $(x,o_1,a_1,o_2)$ for generating a new second response. Its original action $a_2$ is not used as the token-level distillation target. We use $G\geq2$ responses per prefix, and none of their actions is executed.

This sampling scheme follows the prefix-replay setting~\citep{liao2026reopd}: responses are sampled from the current rollout policy, while histories remain distributed according to $q_{\mathcal B}$. It therefore provides on-policy response generation conditional on offline histories, without claiming on-policy coverage of the student's environment-induced histories. The resulting behavior-prefix shift motivates controlling the learning signals assigned to responses within each recorded context.

For a response $y_i=(y_{i,1},\ldots,y_{i,L_i})$, define $c_{i,k}=(h,y_{i,<k})$. Here $L_i$ counts tokens in one response, rather than a complete multi-turn trajectory. The fixed teacher $\pi_T$, assumed to share the student's tokenization, scores these same student-generated tokens. We use the detached teacher--student signal and token importance ratio
\begin{equation}
    \begin{aligned}
    d_{i,k}&=\operatorname{sg}\!\left[
    \log\pi_T(y_{i,k}\mid c_{i,k})-\log\pi_{\mathrm{old}}(y_{i,k}\mid c_{i,k})\right],\\
    \rho_{i,k}(\theta)&=\frac{\pi_\theta(y_{i,k}\mid c_{i,k})}
    {\pi_{\mathrm{old}}(y_{i,k}\mid c_{i,k})}.
    \end{aligned}
    \label{eq:sapd-token-signal}
\end{equation}
where $\operatorname{sg}$ denotes stop-gradient. The sampled token-level OPD surrogate is
\begin{equation}
    \mathcal L_{\mathrm{tok}}(\theta)
    =-\mathbb E\!\left[
    \frac{1}{G}\sum_{i=1}^{G}\frac{1}{L_i}\sum_{k=1}^{L_i}
    \rho_{i,k}(\theta)d_{i,k}\right].
    \label{eq:sapd-token-loss}
\end{equation}
All expectations in this section follow Equation~\ref{eq:prefix-sampling}; sampled responses, teacher scores, and rollout-policy probabilities are held fixed during an update. This token-local surrogate uses the negative sampled reverse-KL signal introduced in the background, covering both reasoning and action tokens. Prefix tokens receive no training loss.

\subsection{Step-Advantage Policy Distillation}
\label{sec:step-advantage}

Token-level distillation supplies local learning directions, but does not explicitly compare complete decisions made under the same history. We introduce a complementary step-level signal by comparing the teacher--student scores of responses within each prefix-conditioned group. In this section, a \emph{step} denotes one agent decision turn, not a token or an optimization iteration.

\paragraph{Action-based step scores.}
Let $\mathcal A_i\subseteq\{1,\ldots,L_i\}$ index the action tokens in response $y_i$, excluding its reasoning segment. We aggregate the teacher--student signal over this set:
\begin{equation}
    r_i^{\mathrm{step}}=\sum_{k\in\mathcal A_i}d_{i,k},
    \qquad
    \bar r^{\mathrm{step}}(h)=\frac{1}{G}\sum_{j=1}^{G}r_j^{\mathrm{step}},
    \qquad
    A_i^{\mathrm{step}}=\operatorname{sg}\!\left[
    r_i^{\mathrm{step}}-\bar r^{\mathrm{step}}(h)\right].
    \label{eq:sapd-step-advantage}
\end{equation}
Restricting the score to action tokens focuses the comparison on the decision and prevents reasoning length from directly contributing to the aggregate. This choice also limits length-dependent scoring error: as Equation~\ref{eq:token-to-step-error} shows, a shared per-token bias leaves a residual proportional to differences in the number of scored tokens. Excluding reasoning tokens avoids an additional source of such variation from potentially long and variable reasoning traces. Reasoning still conditions the teacher and student probabilities and receives token-level supervision through Equation~\ref{eq:sapd-token-loss}. Action-length differences can still contribute residual error; the sum is not length normalized. A positive advantage indicates an above-group-average teacher-to-student log-likelihood ratio, rather than verified environment success.

\paragraph{Relative credit under prefix shift.}
In online multi-turn OPD, student-generated histories can be unfamiliar to the teacher, making its supervision unreliable; a large teacher--student gap can exacerbate this mismatch~\citep{liao2026reopd}. With heterogeneous offline data, a recorded prefix can additionally be atypical under both policies. We therefore seek tolerance to context-dependent scoring errors, rather than assuming that every teacher likelihood is well calibrated on the offline pool.

We begin with the token-level signal in Equation~\ref{eq:sapd-token-signal}. For a fixed prefix $h$ and its sampled responses, let $d_{i,k}^\star$ denote an ideal calibrated supervisory signal, defined formally in Appendix~\ref{app:step-error-model}. On the scored action tokens, consider the structured error model
\begin{equation}
    d_{i,k}=d_{i,k}^\star+b_{\mathrm{tok}}(h)+\xi_{i,k},
    \qquad k\in\mathcal A_i,
    \label{eq:token-error-model}
\end{equation}
where $b_{\mathrm{tok}}(h)$ is a prefix-dependent component shared across the scored tokens and $\xi_{i,k}$ captures token- and response-dependent deviations. This is an assumption about sampled scoring errors, not a property guaranteed by OOD inputs. With $N_i=|\mathcal A_i|$, $r_i^\star=\sum_{k\in\mathcal A_i}d_{i,k}^\star$, and $u_i=\sum_{k\in\mathcal A_i}\xi_{i,k}$, aggregation and group centering give
\begin{equation}
    \begin{aligned}
    r_i^{\mathrm{step}}-r_i^\star&=N_i b_{\mathrm{tok}}(h)+u_i,\\
    A_i^{\mathrm{step}}-A_i^\star
    &=b_{\mathrm{tok}}(h)(N_i-\bar N)+(u_i-\bar u),
    \end{aligned}
    \label{eq:token-to-step-error}
\end{equation}
where bars denote group means and $A_i^\star=r_i^\star-G^{-1}\sum_jr_j^\star$. For equal action lengths, the shared token offset cancels exactly, leaving only differences between accumulated residual errors. With variable lengths, a length-dependent term remains. Thus, the benefit comes from removing the shared component of the aggregate error, not from assuming that summation makes every token error disappear.

More generally, write the aggregate error as a shared offset plus a residual:
\begin{equation}
    r_i^{\mathrm{step}}=r_i^\star+b(h)+e_i,
    \qquad
    A_i^{\mathrm{step}}-A_i^\star=e_i-\bar e,
    \label{eq:step-error-decomposition}
\end{equation}
where $\bar e=G^{-1}\sum_je_j$. For fixed action length $N$, Equation~\ref{eq:token-error-model} yields $b(h)=N b_{\mathrm{tok}}(h)$ and $e_i=u_i$; otherwise, $e_i$ must also account for length-dependent error. A useful robustness guarantee requires these residual differences to be small.

\noindent\textbf{Proposition 1 (Robustness to shared score offsets).}
Let $P=I_G-G^{-1}\mathbf 1\mathbf 1^\top$ be the group-centering matrix, with $I_G$ the identity and $\mathbf 1$ the all-ones vector. For every fixed group, the advantage error is $Pe$, independent of $b(h)$, and
\begin{equation}
    \|A^{\mathrm{step}}-A^\star\|_2^2
    =\|Pe\|_2^2
    =\sum_{i=1}^{G}(e_i-\bar e)^2
    \leq\sum_{i=1}^{G}e_i^2.
    \label{eq:step-projection-bound}
\end{equation}
Moreover, if $|e_i|\leq\varepsilon_h$, then $|A_i^{\mathrm{step}}-A_i^\star|\leq2\varepsilon_h$; an ideal advantage with magnitude greater than $2\varepsilon_h$ retains its sign. Here $\varepsilon_h$ bounds scoring error and is distinct from the clipping threshold $\epsilon$. Thus, relative step credit can remain informative despite large shared offsets, provided response-dependent errors are small. Appendix~\ref{app:step-robustness} proves the result, derives a local update-error bound, and discusses its limitations. This property does not correct arbitrary teacher errors or restore missing coverage of student-relevant histories.

\paragraph{Relative credit without variance normalization.}
We retain the magnitude of the centered signal instead of dividing by its group standard deviation. For $r=b\mathbf 1+\alpha v$, where $\alpha\geq0$ controls the size of within-group differences, the advantage is $Pr=\alpha Pv$. It vanishes as these differences vanish. By comparison, standardization gives $Pr/(\operatorname{std}(r)+\eta)=\alpha Pv/(\alpha\operatorname{std}(v)+\eta)$: when the stabilizer $\eta$ is negligible, even small nonzero differences can produce an order-one signal. Omitting this division avoids such amplification, although a small score spread alone does not establish teacher reliability.

\paragraph{Step-level policy updates.}
We broadcast $A_i^{\mathrm{step}}$ to the generated tokens of response $i$ and use the clipped surrogate
\begin{equation}
    \begin{aligned}
    \mathcal C_\epsilon(\rho,A)
    &=\min\!\left\{\rho A,
    \operatorname{clip}(\rho,1-\epsilon,1+\epsilon)A\right\},\\
    \mathcal L_{\mathrm{step}}(\theta)
    &=-\mathbb E\!\left[\frac{1}{G}\sum_{i=1}^{G}\frac{1}{L_i}
    \sum_{k=1}^{L_i}\mathcal C_\epsilon\!\left(\rho_{i,k}(\theta),A_i^{\mathrm{step}}\right)\right].
    \end{aligned}
    \label{eq:sapd-step-loss}
\end{equation}
The scoring mask and update mask serve different purposes: only action tokens determine $r_i^{\mathrm{step}}$, whereas the response-level advantage updates both reasoning and action tokens. This supplies a shared decision-level signal alongside the distinct per-token signals in $\mathcal L_{\mathrm{tok}}$.

\subsection{Joint Training Objective}
\label{sec:joint-objective}

Optimizing likelihood-based supervision can create opportunities for supervision hacking: the student may exploit favorable teacher scores for degenerate responses without improving its decisions. For example, repeatedly issuing the same action can become a spurious local optimum of the training surrogate. Group-relative step advantages compare responses using the same likelihood-based signal and therefore do not explicitly rule out such behavior. We therefore introduce locally computable rule rewards to discourage these failure modes, including penalties for repeated actions detectable from the recorded history and checks on the generated action format. These signals constrain exploitable response patterns independently of teacher likelihoods and require no environment execution. Let $f_m(h,y_i)$ be the score of rule $m$ and $\alpha_m\geq0$ its weight, with violations represented by negative scores. Define
\begin{equation}
    r_i^{\mathrm{rule}}=\sum_{m=1}^{M}\alpha_m f_m(h,y_i),
    \qquad
    A_i^{\mathrm{rule}}=\operatorname{sg}\!\left[
    \frac{r_i^{\mathrm{rule}}-\bar r^{\mathrm{rule}}}
    {\operatorname{std}(r_1^{\mathrm{rule}},\ldots,r_G^{\mathrm{rule}})+\eta}\right],
    \label{eq:sapd-rule-advantage}
\end{equation}
where $M$ is the number of rules and $\bar r^{\mathrm{rule}}$ is their group-mean aggregate reward. The rule objective uses the same clipped surrogate:
\begin{equation}
    \begin{aligned}
    \mathcal L_{\mathrm{rule}}(\theta)
    &=-\mathbb E\!\left[\frac{1}{G}\sum_{i=1}^{G}\frac{1}{L_i}
    \sum_{k=1}^{L_i}\mathcal C_\epsilon\!\left(\rho_{i,k}(\theta),A_i^{\mathrm{rule}}\right)\right],\\
    \mathcal L_{\mathrm{SAPD}}
    &=\lambda_{\mathrm{tok}}\mathcal L_{\mathrm{tok}}
    +\lambda_{\mathrm{step}}\mathcal L_{\mathrm{step}}
    +\lambda_{\mathrm{rule}}\mathcal L_{\mathrm{rule}}.
    \end{aligned}
    \label{eq:sapd-joint-loss}
\end{equation}
The nonnegative coefficients $\lambda_{\mathrm{tok}}$, $\lambda_{\mathrm{step}}$, and $\lambda_{\mathrm{rule}}$ control the contributions of the three objectives. As shown by the ablation experiments presented later, the performance gains primarily stem from step advantages, while auxiliary token-level OPD improves learning efficiency. Rule rewards further discourage supervision hacking.

Rule definitions and coefficient values are specified in the experimental setup. Each iteration samples offline prefixes, generates response groups, computes teacher and rule scores, and minimizes the joint loss with respect to the student alone. The rollout policy is refreshed from the student for subsequent sampling. All supervision is computed from recorded histories and model-generated responses, with no action execution, new environment observation, or environment-return evaluation during policy refinement.

\section{Experiments}
\label{sec:experiments}

\subsection{Experimental Setup}
\label{sec:experimental-setup}

\textbf{Benchmarks.} We evaluate SAPD on three widely used long-horizon agent tasks: ALFWorld, WebShop, and Search-based QA. \textbf{ALFWorld}~\citep{shridhar2021alfworld} is a text-based environment in which agents follow instructions to complete household goals. It comprises six task categories: Pick, Look, Clean, Heat, Cool, and Pick2. \textbf{WebShop}~\citep{yao2022webshop} is a simulated online shopping environment in which agents search a product catalog, navigate product pages, and select products and options that satisfy user-specified requirements. \textbf{Search-based QA} follows the setup of Search-R1~\citep{jin2025search}, requiring agents to interleave reasoning with search, gather relevant evidence, and produce a final answer. It includes three single-hop datasets, Natural Questions~\citep{kwiatkowski2019natural}, TriviaQA~\citep{joshi2017triviaqa}, and PopQA~\citep{mallen2023not}, and four multi-hop datasets, HotpotQA~\citep{yang2018hotpotqa}, 2WikiMultihopQA~\citep{ho2020constructing}, MuSiQue~\citep{trivedi2022musique}, and Bamboogle~\citep{press2023measuring}. Following the setup of SDAR~\citep{lu2026self}, we evaluate 128 episodes per evaluation round on each of ALFWorld and WebShop, and all 51,713 questions across the seven Search-based QA evaluation datasets.

\textbf{Baselines.} To evaluate the effectiveness of SAPD, we compare against the following three categories of baselines. \textbf{(1) Supervised Fine-Tuning (SFT)}~\citep{zeng2024agenttuning} learns to imitate expert actions in fixed demonstration trajectories. As a fully off-policy baseline, SFT allows us to assess SAPD's task performance and resistance to catastrophic forgetting under sequential task training. \textbf{(2) Group Relative Policy Optimization (GRPO)}~\citep{shao2024deepseekmath} collects trajectories by having the student interact with the environment and updates the policy using advantages computed from relative rewards within each rollout group. Its reliance on extensive environment interaction provides a comparison for SAPD in both task performance and training efficiency. \textbf{(3) On-Policy Distillation (OPD)}~\citep{agarwal2024onpolicy} applies token-level teacher supervision to student-generated responses conditioned on offline SFT prefixes. OPD+Rule further incorporates rule-based rewards to stabilize training. Its objective differs from SAPD only by omitting our step-advantage policy distillation, isolating the contribution of action-level guidance.

\textbf{Implementation Details.} We use Qwen2.5-1.5B-Instruct as the student backbone and SFT- or GRPO-trained Qwen2.5-3B-Instruct models as frozen teachers~\citep{qwen2025qwen25technicalreport}. The student samples eight responses per decision prefix extracted from offline SFT trajectories without additional environment interaction. The training framework and experimental hyperparameters follow SDAR~\citep{lu2026self}. Full details of implementation are provided in Appendix~\ref{app:experimental-details}.

\subsection{Main Results}
\label{sec:main-results}

Table~\ref{tab:main-results} shows the
performance of SAPD across two teacher configurations and three long-horizon agentic benchmarks.
Based on these experiments, we summarize the following key observations:

\textbf{SAPD improves upon OPD-based methods.}
Table~\ref{tab:main-results} demonstrates the effectiveness of SAPD in both benchmark performance and learning dynamics. SAPD outperforms OPD+Rule on all three benchmarks under both teacher configurations. With GRPO teachers, the gains are 6.5, 0.3, and 4.4 percentage points on ALFWorld, Search-based QA, and WebShop, respectively; with SFT teachers, the corresponding improvements are 2.6, 0.2, and 6.3 points.
These gains highlight SAPD's ability to mitigate distillation instability in long-horizon tasks, where the teacher may provide unreliable supervision on student-induced trajectories due to distribution mismatch.

\textbf{Teacher strength alone does not determine distillation quality.}
On WebShop, the GRPO teacher achieves a success rate of 64.8\%, substantially exceeding the SFT teacher's 59.7\%. However, the student trained with the SFT teacher reaches 59.4\%, compared with 53.4\% for the GRPO teacher.
This difference highlights the distinction between a teacher's standalone performance and its effectiveness as a supervisor.
In our experimental setting, all distillation methods condition student generation on prefixes extracted from offline SFT trajectories, and teachers provide feedback on the resulting
student-generated responses.
These prefixes reflect SFT behavior and may differ from the histories encountered by the GRPO teacher under its
own policy. Consequently, stronger performance during autonomous execution does not guarantee more useful supervision in the distillation setting.
The WebShop results therefore show that teacher selection should account for compatibility with the distillation contexts, rather than standalone task performance alone.

\textbf{Long-horizon tasks benefit more from SAPD.}
SAPD provides larger gains on ALFWorld and WebShop than on Search-based QA, as shown in Table~\ref{tab:main-results}. This difference can be understood in terms of interaction
horizon: ALFWorld allows up to 50 steps and WebShop up to 15, compared with four for Search-based QA.
Over longer interactions, early decisions shape subsequent observations and actions, allowing errors and mismatches between teacher and student behavior to accumulate.
SAPD compares candidate actions under the same prefix, reducing the impact of context mismatch on the distillation signal.
The resulting relative step-level advantages provide more
stable guidance for policy optimization, contributing to
larger gains on longer-horizon tasks.


\begin{table*}[t]
\centering
\begingroup
\small
\setlength{\tabcolsep}{3.5pt}
\renewcommand{\arraystretch}{1.12}

\resizebox{1.0\textwidth}{!}{%
\begin{tabular}{@{}l*{15}{r}c@{}}
\toprule[0.9pt]
 & \multicolumn{7}{c}{\textbf{ALFWorld}}
 & \multicolumn{8}{c}{\textbf{Search-based QA}}
 & \textbf{WebShop} \\
\cmidrule(lr){2-8}
\cmidrule(lr){9-16}
\cmidrule(l){17-17}
\textbf{Method}
 & Pick & Look & Clean & Heat & Cool & Pick2 & \textbf{Avg.}
 & NQ & Triv & Pop & Hotp & 2Wk & MuS & Bam & \textbf{Avg.}
 & \textbf{Success} \\
\midrule

SFT
 & 100.0 & 100.0 & 94.1 & 73.7 & 88.2 & 84.2 & 90.6
 & 32.8 & 47.4 & 38.3 & 29.9 & 36.5 & 11.1 & 28.0 & 37.0
 & 51.6 \\

GRPO
 & 85.3 & 53.7 & 84.5 & 78.2 & 59.7 & 53.5 & 72.8
 & 34.1 & 54.5 & 37.8 & 32.4 & 31.9 & 10.3 & 26.4 & 37.7
 & 56.8 \\

\midrule[0.35pt]\midrule[0.35pt]
\multicolumn{17}{@{}l@{}}{
    \textit{GRPO teacher model: Qwen2.5-3B-Instruct}
} \\
\addlinespace[2pt]

GRPO
 & 96.8 & 88.2 & 92.9 & 91.7 & 90.0 & 70.0 & 89.1
 & 43.4 & 60.8 & 46.2 & 39.3 & 35.3 & 14.8 & 66.9 & 44.5
 & 64.8 \\

\midrule[0.35pt]

OPD
 & 83.4 & 60.7 & 72.7 & 49.1 & 42.0 & 13.3 & 57.0
 & 11.1 & 25.6 & 16.9 & 8.6 & 15.4 & 1.7 & 12.8 & 16.1
 & 29.9 \\

    OPD+Rule
 & 83.0 & 48.0 & 82.5 & 72.5 & 55.1 & 16.5 & 62.8
 & \textbf{40.8} & 55.4 & 42.9 & 36.2
 & 34.3 & 12.7 & \textbf{31.2} & 41.0
 & 49.0 \\

\addlinespace[2pt]
\rowcolor{gray!10}
\textbf{SAPD}
 & \textbf{90.7} & \textbf{60.9} & \textbf{82.7}
 & \textbf{83.7} & \textbf{72.9} & \textbf{18.9} & \textbf{69.3}
 & 40.0 & \textbf{56.1} & \textbf{43.2} & \textbf{36.5}
 & \textbf{34.5} & \textbf{13.3} & 28.8 & \textbf{41.3}
 & \textbf{53.4} \\

\midrule[0.35pt]\midrule[0.35pt]
\multicolumn{17}{@{}l@{}}{
    \textit{SFT teacher model: Qwen2.5-3B-Instruct}
} \\
\addlinespace[2pt]

SFT
 & 100.0 & 100.0 & 93.5 & 89.5 & 91.3 & 95.0 & 94.5
 & 40.7 & 58.4 & 43.0 & 36.5 & 34.6 & 12.5 & 63.3 & 42.0
 & 59.7 \\

\midrule[0.35pt]

OPD
 & 90.3 & \textbf{93.0} & 87.0 & 84.0 & 73.2 & 85.9 & 86.7
 & 9.0 & 22.6 & 12.7 & 6.3 & 7.2 & 2.2 & 6.4 & 11.9
 & 47.1 \\

OPD+Rule
 & 97.7 & 82.9 & \textbf{95.4} & 90.5 & 96.8 & 95.5 & 94.0
 & \textbf{34.1} & 49.4 & 39.9 & \textbf{35.9}
 & 35.1 & \textbf{15.0} & 32.8 & 38.6
 & 53.1 \\

\addlinespace[2pt]
\rowcolor{gray!10}
\textbf{SAPD}
 & \textbf{99.0} & 85.9 & 94.5 & \textbf{94.8}
 & \textbf{100.0} & \textbf{98.1} & \textbf{96.6}
 & 34.0 & \textbf{49.5} & \textbf{40.1} & 35.4
 & \textbf{36.1} & 14.1 & \textbf{37.6} & \textbf{38.8}
 & \textbf{59.4} \\

\bottomrule[0.9pt]
\end{tabular}%
}
\endgroup
\caption{Main results on ALFWorld, Search-based QA,
and WebShop with SFT and GRPO teachers.
We report success rates on ALFWorld and WebShop and
accuracy on Search-based QA.
The best results are bolded, and the second-best results are underlined.}
\label{tab:main-results}
\end{table*}

\begin{table*}[htbp]
\centering

\begin{minipage}[t]{0.44\textwidth}
\vspace{0pt}
\centering
\small
\setlength{\tabcolsep}{3pt}
\renewcommand{\arraystretch}{1.0}

\begin{tabular}{@{}lccc@{}}
\toprule[0.8pt]
& \textbf{ALFWorld}
& \textbf{Search}
& \textbf{WebShop} \\
\cmidrule(lr){2-4}
\textbf{Method}
& Success
& Overall
& Success \\
\midrule

\multicolumn{4}{@{}l@{}}{\textit{GRPO teacher}} \\
\addlinespace[1pt]

\rowcolor{gray!10}
\textbf{SAPD}
& \textbf{69.3} & \textbf{41.3} & \textbf{53.4} \\
\quad w/o OPD
& 66.9 & 41.0 & 51.6 \\
\quad w/o Rule
& 62.0 & 40.7 & 52.1 \\

\midrule[0.35pt]\midrule[0.35pt]
\multicolumn{4}{@{}l@{}}{\textit{SFT teacher}} \\
\addlinespace[1pt]

\rowcolor{gray!10}
\textbf{SAPD}
& \textbf{96.6} & \textbf{38.8} & \textbf{59.4} \\
\quad w/o OPD
& 95.8 & 38.7 & 56.0 \\
\quad w/o Rule
& 95.3 & 38.3 & 55.2 \\

\bottomrule[0.8pt]
\end{tabular}

\caption{Results of the ablation study.}
\label{tab:ablation}
\end{minipage}\hfill
%
\begin{minipage}[t]{0.54\textwidth}
\vspace{0pt}
\centering
\small
\setlength{\tabcolsep}{3pt}
\renewcommand{\arraystretch}{1.0}

\begin{tabular}{@{}lccc@{}}
\toprule[0.8pt]
& \multicolumn{3}{c}{\textbf{Evaluation after training}} \\
\cmidrule(lr){2-4}
\textbf{Training stage}
& \textbf{WebShop}
& \textbf{ALFWorld}
& \textbf{Search} \\
\midrule

\multicolumn{4}{@{}l@{}}{\textbf{SAPD}} \\
\addlinespace[1pt]

Stage 1: WebShop
& \cellcolor{SAPDRetain}\textbf{52.3}
& 16.4
& 16.2 \\

Stage 2: ALFWorld
& \cellcolor{SAPDRetain}57.0
& \cellcolor{SAPDRetain}\textbf{96.1}
& 13.9 \\

Stage 3: Search
& \cellcolor{SAPDRetain}50.8
& \cellcolor{SAPDRetain}90.5
& \cellcolor{SAPDRetain}\textbf{36.6} \\

\midrule[0.35pt]\midrule[0.35pt]
\multicolumn{4}{@{}l@{}}{\textbf{SFT}} \\
\addlinespace[1pt]

Stage 1: WebShop
& \cellcolor{SAPDRetain}\textbf{51.6}
& 15.3
& 16.9 \\

Stage 2: ALFWorld
& \cellcolor{SFTRetain}22.7
& \cellcolor{SAPDRetain}\textbf{92.2}
& 10.2 \\

Stage 3: Search
& \cellcolor{SFTRetain}2.3
& \cellcolor{SFTRetain}7.8
& \cellcolor{SAPDRetain}\textbf{37.2} \\

\bottomrule[0.8pt]
\end{tabular}

\caption{Performance under sequential training.}
\label{tab:multistage-comparison}
\end{minipage}

\end{table*}

\subsection{Ablation Study}
Table~\ref{tab:ablation} presents the ablation results for SAPD,
from which we draw two key observations.

\textbf{Step-level advantages provide the primary learning signal,
while token-level OPD plays an auxiliary role.}
To assess the contribution of step-advantage policy optimization, we remove the token-level OPD branch while retaining step-level advantages and rule-based rewards, yielding a variant denoted as w/o OPD.
The w/o OPD variant exhibits only modest performance declines relative to SAPD.
Compared with OPD+Rule in Table~\ref{tab:main-results}, it outperforms the baseline across all benchmarks
and teacher configurations. Specifically, it improves ALFWorld performance by 4.1 points
with a GRPO teacher and 1.8 points with an SFT teacher; the corresponding WebShop gains are 2.6 and 2.9 points.
These results demonstrate that step-advantage policy optimization accounts for most of gains, while
token-level OPD provides dense supervision for further performance improvements.

\textbf{Rule-based rewards help stabilize the training.}
To discourage undesirable output behaviors, we supplement teacher-derived supervision with rule-based rewards.
We isolate their contribution through w/o Rule, which removes these rewards while retaining step-level advantages and token-level OPD.
Removing rule-based rewards consistently degrades performance across all evaluated configurations.
For example, ALFWorld success falls from 69.3\% to 62.0\% with a GRPO teacher, while WebShop success declines from 59.4\% to 55.2\% with an SFT teacher.
These results suggest that rule-based rewards complement teacher-derived learning signals.
Meanwhile, SAPD retains substantial task performance even without rule-based rewards, indicating its effectiveness.

\subsection{Sequential Training Compared With SFT}

Sequential training requires agents to learn new tasks while maintaining performance on earlier tasks. We investigate whether SAPD mitigates catastrophic forgetting relative to SFT by training both methods sequentially on WebShop, ALFWorld, and Search-based QA. In Table~\ref{tab:multistage-comparison}, each row corresponds to a completed training stage. Diagonal entries show performance on the newly learned task, and entries below the diagonal track performance on earlier tasks. The two methods begin with similar WebShop success rates of 52.3\% for SAPD and 51.6\% for SFT, but their performance diverges during subsequent training. Following ALFWorld training, SAPD's WebShop success rises to 57.0\%, possibly because the shared output format supports generalization across tasks. In contrast, SFT's WebShop success falls to 22.7\%, indicating substantial forgetting of the previously learned task. By the end of Search training, SAPD still achieves 50.8\% on WebShop and 90.5\% on ALFWorld, whereas SFT reaches only 2.3\% and 7.8\%, respectively. This retention advantage accompanies comparable accuracy on the final Search task, which demonstrates that SAPD substantially reduces catastrophic forgetting without compromising its ability to learn new tasks.

\begin{figure*}[htbp]
    \centering
    \includegraphics[width=1.0\textwidth]
    {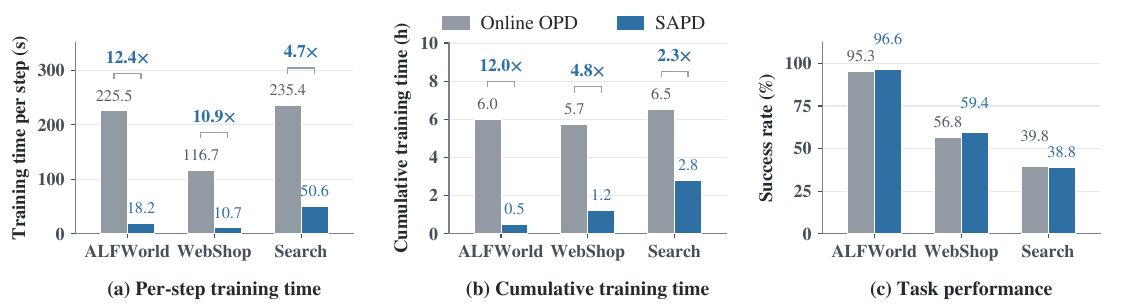}
    \caption{Training efficiency and evaluation performance
    of SAPD and Online OPD.}
    \label{fig:efficiency-effectiveness}
\end{figure*}

\subsection{Training Efficiency Compared with Online OPD}

Repeated environment interaction is a major source of overhead in online agent training, motivating an efficiency comparison between SAPD and online OPD. Figure~\ref{fig:efficiency-effectiveness} compares per-step training time, cumulative training time, and task performance. Compared with online OPD, SAPD achieves per-step speedups of $12.4\times$ on ALFWorld, $10.9\times$ on WebShop, and $4.7\times$ on Search-based QA. These gains also translate into cumulative training-time reductions by factors of approximately $12.0$, $4.8$, and $2.3$, respectively. SAPD achieves these savings with higher performance on ALFWorld and WebShop and slightly lower accuracy on Search-based QA. Furthermore, the observed speedup increases across benchmarks with longer interaction horizons, from Search-based QA with at most four steps to WebShop with 15 and ALFWorld with 50. Longer rollouts require online OPD to repeatedly generate responses and interact with the environment, with repetitive or unsuccessful actions potentially further extending these time costs. SAPD avoids this overhead by reusing offline trajectories, yielding greater efficiency gains on longer-horizon tasks.

\section{Conclusion}
\label{con}

We presented SAPD, a method for refining LLM agents from recorded interaction histories without further environment interaction.
SAPD decomposes multi-turn trajectories into decision prefixes and compares student-generated responses under each shared prefix to construct relative step advantages, providing decision-level guidance alongside token-level distillation.
Experiments on ALFWorld, WebShop, and Search-based QA show improved performance over OPD-based baselines across SFT and GRPO teacher configurations, reduced catastrophic forgetting relative to sequential SFT, and at least a $4.7\times$ reduction in per-step training time compared with online OPD under the evaluated configurations.
These results demonstrate the potential of reusing recorded interaction experience to expand agent capabilities while reducing reliance on executable training environments.





\bibliography{conference}
\bibliographystyle{conference}

\appendix
\section{Analysis of Relative Step Credit under Prefix Shift}
\label{app:step-robustness}

This appendix analyzes sensitivity of the step-advantage branch to teacher scoring errors on a fixed set of offline contexts. The results concern the centered supervision and its local policy update; they are not convergence guarantees or guarantees of improved task success under arbitrary distribution shift.

\subsection{Error Model and the Role of Sequence Aggregation}
\label{app:step-error-model}

Fix a prefix $h$ and $G$ responses sampled according to Equation~\ref{eq:prefix-sampling}. For analysis only, introduce an ideal supervisory distribution $\pi_\star$ that shares the tokenization and assigns positive probability to the scored tokens. The ideal token signal and action score are
\begin{equation}
    d_{i,k}^\star
    =\log\pi_\star(y_{i,k}\mid c_{i,k})
      -\log\pi_{\mathrm{old}}(y_{i,k}\mid c_{i,k}),
    \qquad
    r_i^\star=\sum_{k\in\mathcal A_i}d_{i,k}^\star.
    \label{eq:ideal-step-score}
\end{equation}
The distribution $\pi_\star$ is a theoretical comparator, not an additional model available to SAPD. No equality between $r_i^\star$ and environment return is assumed. The aggregate teacher error is exactly
\begin{equation}
    \Delta_i=r_i^{\mathrm{step}}-r_i^\star
    =\sum_{k\in\mathcal A_i}
    \log\frac{\pi_T(y_{i,k}\mid c_{i,k})}{\pi_\star(y_{i,k}\mid c_{i,k})}.
    \label{eq:aggregate-teacher-error}
\end{equation}
Our structured-error model is $\Delta_i=b(h)+e_i$. The algebra below holds for any such decomposition; a substantive robustness interpretation requires the residual differences $e_i-\bar e$ to be small. In particular, defining $b(h)$ as a mean does not by itself establish that the remaining errors are small. The model permits a teacher to misestimate absolute scores on an unfamiliar prefix while preserving useful relative distinctions among responses.

At the token level, Equation~\ref{eq:token-error-model} decomposes the log-ratio error $d_{i,k}-d_{i,k}^\star=\log\pi_T(y_{i,k}\mid c_{i,k})-\log\pi_\star(y_{i,k}\mid c_{i,k})$ into $b_{\mathrm{tok}}(h)+\xi_{i,k}$. Both policies remain normalized distributions; the decomposition concerns errors on sampled tokens and does not posit an arbitrary additive shift to all log-probabilities in the vocabulary. Summing over the action mask gives
\begin{equation}
    \Delta_i=N_i b_{\mathrm{tok}}(h)+u_i,
    \qquad
    \bar\Delta=\bar N b_{\mathrm{tok}}(h)+\bar u,
    \qquad
    \Delta_i-\bar\Delta
    =b_{\mathrm{tok}}(h)(N_i-\bar N)+u_i-\bar u,
    \label{eq:token-error-centering-proof}
\end{equation}
which proves Equation~\ref{eq:token-to-step-error}. This also connects the token model to the aggregate model: for any fixed reference length $N_0(h)$, choose $b(h)=N_0(h)b_{\mathrm{tok}}(h)$ and $e_i=(N_i-N_0(h))b_{\mathrm{tok}}(h)+u_i$. Proposition~1 then applies without assuming equal lengths, but its residual bound includes length variation.

If $|\xi_{i,k}|\leq\varepsilon_{\mathrm{tok},h}$ on every scored token, then $|u_i|\leq N_i\varepsilon_{\mathrm{tok},h}$ and $|\bar u|\leq\bar N\varepsilon_{\mathrm{tok},h}$. Hence
\begin{equation}
    |A_i^{\mathrm{step}}-A_i^\star|
    \leq |b_{\mathrm{tok}}(h)|\,|N_i-\bar N|
       +(N_i+\bar N)\varepsilon_{\mathrm{tok},h}.
    \label{eq:token-error-advantage-bound}
\end{equation}
For equal lengths $N$, this reduces to $2N\varepsilon_{\mathrm{tok},h}$ independently of the shared token offset. An ideal advantage exceeding the applicable bound in magnitude retains its sign. The bound also makes clear that accumulated residual error can grow with action length under sum aggregation.

Aggregation creates one score per decision to enable these comparisons, but summation alone does not guarantee noise reduction. If token errors $\xi_{i,k}$ are treated as random variables with a fixed action mask, then
\begin{equation}
    \operatorname{Var}\!\left(\sum_{k\in\mathcal A_i}\xi_{i,k}\right)
    =\sum_{k\in\mathcal A_i}\operatorname{Var}(\xi_{i,k})
    +2\!\sum_{\substack{k<\ell\\k,\ell\in\mathcal A_i}}
    \operatorname{Cov}(\xi_{i,k},\xi_{i,\ell}).
    \label{eq:aggregation-variance}
\end{equation}
Consequently, neither independent nor correlated token errors can be assumed to disappear through aggregation. Our cancellation result comes from subsequent group centering.

\subsection{Proof of Proposition 1}
\label{app:centering-proof}

Write $r=r^\star+b(h)\mathbf 1+e$. The matrix $P=I_G-G^{-1}\mathbf 1\mathbf 1^\top$ satisfies $P\mathbf 1=0$, $P^\top=P$, and $P^2=P$. Therefore,
\begin{equation}
    A^{\mathrm{step}}=Pr=Pr^\star+Pe=A^\star+Pe.
\end{equation}
Since $P$ is an orthogonal projection, it cannot increase Euclidean norm. More explicitly,
\begin{equation}
    \|Pe\|_2^2
    =\sum_{i=1}^{G}e_i^2-G\bar e^2
    \leq\sum_{i=1}^{G}e_i^2.
\end{equation}
If $|e_i|\leq\varepsilon_h$, then $|\bar e|\leq\varepsilon_h$ and $|e_i-\bar e|\leq2\varepsilon_h$, proving the claimed sign guarantee. Likewise, pairwise score ordering is preserved whenever $|r_i^\star-r_j^\star|>2\varepsilon_h$, since the error in that difference is $e_i-e_j$. These conclusions are independent of the magnitude of $b(h)$.

For a conditional stochastic interpretation, suppose $\mathbb E[e\mid h]=0$ and $\operatorname{Cov}(e\mid h)=\Sigma_h$. With randomness taken over response groups and their residual errors,
\begin{equation}
    \frac{1}{G}\mathbb E[\|A^{\mathrm{step}}-A^\star\|_2^2\mid h]
    =\frac{1}{G}\operatorname{tr}(P\Sigma_hP)
    =\frac{1}{G}\operatorname{tr}(\Sigma_h)
       -\frac{1}{G^2}\mathbf 1^\top\Sigma_h\mathbf 1.
    \label{eq:relative-score-mse}
\end{equation}
For conditionally uncorrelated residuals with variance $\sigma_h^2$, this becomes $(1-G^{-1})\sigma_h^2$, whereas the uncentered score error has mean squared magnitude $b(h)^2+\sigma_h^2$. This compares observed and ideal versions of each score representation, not the task performance of two algorithms. It shows that improvement can be substantial when shared offsets dominate, but that centering does not average away arbitrary response-dependent noise.

\subsection{Effect on the Local Step-Advantage Update}
\label{app:step-update-bound}

Consider the first gradient evaluation at $\theta=\theta_{\mathrm{old}}$, before any clipping is active. Define the length-normalized response score gradient
\begin{equation}
    z_i=\left.\frac{1}{L_i}\sum_{k=1}^{L_i}
    \nabla_\theta\log\pi_\theta(y_{i,k}\mid c_{i,k})
    \right|_{\theta=\theta_{\mathrm{old}}},
    \qquad \|z_i\|_2\leq M_h.
\end{equation}
Here $M_h$ is a bound for the fixed group. With detached advantages, the ascent direction associated with $-\mathcal L_{\mathrm{step}}$ is $g=G^{-1}\sum_i A_i^{\mathrm{step}}z_i$. Let $g^\star$ use the ideal centered advantages on the same responses. Cauchy--Schwarz gives
\begin{equation}
    \begin{aligned}
    \|g-g^\star\|_2
    &=\left\|\frac{1}{G}\sum_{i=1}^{G}(Pe)_iz_i\right\|_2\\
    &\leq\frac{1}{G}\|Pe\|_2
        \left(\sum_{i=1}^{G}\|z_i\|_2^2\right)^{1/2}
     \leq\frac{M_h}{\sqrt G}\|Pe\|_2.
    \end{aligned}
    \label{eq:step-update-error}
\end{equation}
Thus shared offsets do not perturb this local step update; residual scoring differences control its deviation from the ideal comparator. If $|e_i|\leq\varepsilon_h$, projection contraction further implies $\|g-g^\star\|_2\leq M_h\varepsilon_h$. Under the uncorrelated residual model above and a bound $M_h$ uniform over sampled groups, the conditional mean squared update error is at most $M_h^2(1-G^{-1})\sigma_h^2$. This is a local sensitivity statement, not a global optimization result. We do not assume that the length-normalized, group-centered estimator is an unbiased policy gradient.

Shared-offset invariance also holds exactly for the clipped loss in Equation~\ref{eq:sapd-step-loss} at any fixed $\theta$, because adding $b(h)$ does not change its advantages. For a bound on loss perturbations, let $K$ bound both $\rho_{i,k}(\theta)$ and $\operatorname{clip}(\rho_{i,k}(\theta),1-\epsilon,1+\epsilon)$ on the sampled tokens. The function $\mathcal C_\epsilon(\rho,A)$ is $K$-Lipschitz in $A$, so the group loss obeys
\begin{equation}
    |\ell_{\mathrm{step}}(r;\theta)-\ell_{\mathrm{step}}(r^\star;\theta)|
    \leq\frac{K}{G}\sum_{i=1}^{G}|(Pe)_i|
    \leq\frac{K}{\sqrt G}\|Pe\|_2,
\end{equation}
where $\ell_{\mathrm{step}}$ is the per-group summand defining $\mathcal L_{\mathrm{step}}$. This bound does not assert gradient smoothness at clipping boundaries.

\subsection{Why Omit Standard-Deviation Normalization?}
\label{app:no-std}

Let $r=b\mathbf 1+\alpha v$ with $\alpha\geq0$ and $Pv\neq0$. Denote the population standard deviation across the $G$ group entries by $s(v)=\|Pv\|_2/\sqrt G$. Centering and standardization give, respectively,
\begin{equation}
    A=\alpha Pv,
    \qquad
    A_{\mathrm{std}}=\frac{\alpha Pv}{\alpha s(v)+\eta},
    \qquad
    \frac{\|A_{\mathrm{std}}\|_2}{\|A\|_2}
    =\frac{1}{\alpha s(v)+\eta}\quad(\alpha>0).
\end{equation}
Without a stabilizer, $\|A_{\mathrm{std}}\|_2=\sqrt G$ for every $\alpha>0$, even as the unnormalized signal tends to zero. With a fixed $\eta>0$, the normalized signal also tends to zero eventually, but its gain relative to $A$ can approach $1/\eta$. SAPD preserves the original scale of within-group differences, avoiding this potentially large gain. Both versions cancel shared additive offsets; the benefit of omitting normalization is magnitude control, not additional offset cancellation. This argument neither proves that small score spread implies uncertainty nor that normalization is universally harmful.

\subsection{Scope and Empirical Implications}
\label{app:robustness-scope}

Three limitations are material. First, the common-offset assumption is at the aggregate action-score level. As Equation~\ref{eq:token-to-step-error} shows, a shared token error $b_{\mathrm{tok}}(h)$ leaves the residual $b_{\mathrm{tok}}(h)(N_i-\bar N)$ when action lengths vary. Thus, action-token summation does not automatically turn a shared token error into a shared sequence error. This provides an additional motivation for the action-only scoring mask: it excludes reasoning-length variation from the aggregate error. It does not guarantee lower error for every response group, since token biases and residuals can also differ between reasoning and action segments. The bound must still account for action-length variation; fixed-length or length-controlled comparisons can isolate it empirically.

Second, centering does not repair incorrect relative teacher preferences. For example, a teacher that reverses the ranking of two responses also reverses their centered advantages. Nor does centering alter $q_{\mathcal B}$: histories absent from the offline pool remain unavailable, and familiarity to either policy is not guaranteed. Offline replay can introduce an additional mismatch relative to online OPD, but need not be more out of distribution for every dataset or teacher.

Third, the guarantees apply to the step branch, not automatically to the combined objective. Token-level OPD remains sensitive to its own residual errors, and the weights in Equation~\ref{eq:sapd-joint-loss} determine how the branches interact. Baseline terms can also cancel in expectation in conventional score-function estimators, so the projection analysis alone does not establish a universally less biased or lower-variance update than original token-level OPD.

The analysis motivates testing three separate predictions: exact invariance of the step branch to injected groupwise score offsets; sensitivity to response-dependent perturbations and teacher ranking errors; and the effect of standard-deviation normalization under weak within-group score differences. These controls should be complemented by comparisons across trajectory sources at fixed teacher and student checkpoints, with matched local rewards and action-length statistics. Such experiments are needed to establish whether the structured-error regime explains practical tolerance to unfamiliar prefixes.

\section{Experimental Details}
\label{app:experimental-details}
This appendix describes the offline data, training configuration, and rule-based rewards used in our experiments. We build on the training framework of SDAR~\citep{lu2026self}.

\subsection{Offline Data Preparation}
\label{app:offline-data}
\paragraph{Trajectory sources.} For ALFWorld and WebShop, we reuse the offline expert trajectories released with SPA-RL~\citep{wang2025spa}, which contain task instructions, agent responses, and environment observations. For WebShop, success filtering yields 909 training trajectories and 3,384 decision prefixes. Search-based QA combines the Search subset of SkillRL~\citep{xia2026skillrl} and the AFM-MHQA~\citep{li2025chainofagentsendtoendagentfoundation} dataset. Table~\ref{tab:appendix-offline-data} summarizes the offline pools before tokenizer-based length filtering. Each trajectory contributes one prefix per recorded decision.

\begin{table}[htbp]
\centering
\small
\setlength{\tabcolsep}{6pt}
\begin{tabular}{lrrrr}
\toprule
& \multicolumn{2}{c}{Training pool} & \multicolumn{2}{c}{Offline validation pool} \\
\cmidrule(lr){2-3}\cmidrule(lr){4-5}
Benchmark & Trajectories & Prefixes & Trajectories & Prefixes \\
\midrule
ALFWorld & 2,382 & 21,300 & 123 & 1,132 \\
WebShop & 909 & 3,384 & 45 & 166 \\
Search-based QA & 8,146 & 22,826 & 428 & 1,198 \\
\bottomrule
\end{tabular}
\caption{Statistics of the offline trajectory pools. Offline validation pools are distinct from the environment episodes and QA evaluation sets used to measure task performance.}
\label{tab:appendix-offline-data}
\end{table}

\paragraph{Prefix construction.} For each recorded decision, we assemble a prompt containing the task instruction, preceding action--observation pairs, and current observation. The target decision and subsequent observations are excluded from the prompt. We retain the trajectory identifier, decision index, and recorded action as metadata; the trajectory identifier and decision index jointly identify a prefix. Task-specific templates use \texttt{<think>} blocks for reasoning and \texttt{<action>} blocks for ALFWorld and WebShop decisions. Search responses use either a \texttt{<search>} block or an \texttt{<answer>} block, while retrieved passages appear in the recorded history within \texttt{<information>} blocks.

\paragraph{Data partitioning and batching.} We use a nominal 5\% offline validation split with random seed 0. Each training batch includes all retained decision prefixes from a sampled set of trajectories. Accordingly, the batch sizes in Table~\ref{tab:appendix-training} refer to trajectories, while the numbers of decision contexts and generated responses depend on trajectory lengths. Candidate advantages are computed separately within each group sharing the same prefix. The processed offline datasets are provided in the supplementary materials.

\begin{table}[htbp]
\centering
\small
\setlength{\tabcolsep}{5pt}
\begin{tabular}{lccc}
\toprule
Setting & ALFWorld & WebShop & Search-based QA \\
\midrule
Student model & \multicolumn{3}{c}{Qwen2.5-1.5B-Instruct} \\
Teacher model & \multicolumn{3}{c}{Qwen2.5-3B-Instruct} \\
Teacher training & \multicolumn{3}{c}{SFT or GRPO} \\
Teacher updates & \multicolumn{3}{c}{Frozen throughout distillation} \\
\midrule
Learning rate & $10^{-6}$ & $10^{-6}$ & $10^{-6}$ \\
Trajectories per training batch & 16 & 16 & 128 \\
Responses per prefix & 8 & 8 & 8 \\
Maximum prompt tokens & 8,192 & 8,192 & 8,192 \\
Maximum response tokens & 512 & 512 & 512 \\
Training temperature & 1.0 & 1.0 & 1.0 \\
PPO mini-batch size & 256 & 256 & 256 \\
Micro-batch size per GPU & 8 & 8 & 8 \\
PPO epochs per sampled batch & 1 & 1 & 1 \\
Clipping threshold & 0.2 & 0.2 & 0.2 \\
Gradient clipping norm & 1.0 & 1.0 & 1.0 \\
Weight decay & 0.01 & 0.01 & 0.01 \\
Entropy coefficient & 0 & 0 & 0 \\
Default training updates & 300 & 450 & 300 \\
\midrule
Evaluation decoding & Greedy & Greedy & Greedy \\
Evaluation interval (updates) & 5 & 5 & 100 \\
Maximum evaluation turns & 50 & 15 & 4 \\
Episodes/questions per evaluation & 128 & 128 & 51,713 \\
Retrieved passages per search & --- & --- & 3 \\
\bottomrule
\end{tabular}
\caption{Model settings and default training and evaluation configurations.}
\label{tab:appendix-training}
\end{table}

\subsection{Models and Training Configuration}
\label{app:training-config}

We use Qwen2.5-1.5B-Instruct as the student and frozen SFT- or GRPO-trained Qwen2.5-3B-Instruct models as teachers. The SFT teachers are trained on the corresponding offline datasets described in Appendix~\ref{app:offline-data}, which are provided in the supplementary materials. We train the student on 8 NVIDIA H200 GPUs using the SDAR framework~\citep{lu2026self}, with \texttt{vLLM} for generation. The student samples eight responses per offline prefix, and only generated response tokens contribute to the training loss. In Equation~\ref{eq:sapd-joint-loss}, we set $\lambda_{\mathrm{tok}}=0.1$, $\lambda_{\mathrm{step}}=0.5$, and $\lambda_{\mathrm{rule}}=0.1$, with the rule coefficient following SDAR~\citep{lu2026self}. Token-level feedback is additionally scaled by 0.15. Table~\ref{tab:appendix-training} summarizes the default configurations. The efficiency analysis in Figure~\ref{fig:efficiency-effectiveness} reports training time only, excluding evaluation and checkpoint saving for both methods.

\subsection{Rule-Based Rewards}
\label{app:rule-details}
To discourage malformed responses and repetitive actions, we use two locally computable rule checks. For format checking, we verify the required response tags and complete decision blocks: \texttt{<action>} blocks for ALFWorld and WebShop, and \texttt{<search>} or \texttt{<answer>} blocks for Search-based QA. Missing or incomplete required tags receive a negative format score. For repetition checking, we extract the generated action and assign a negative score if it exactly matches the immediately preceding action in the recorded prefix. Rule scores are aggregated and standardized within each prefix group following Equation~\ref{eq:sapd-rule-advantage}. The resulting advantages supervise all generated response tokens through Equation~\ref{eq:sapd-joint-loss}, with $\lambda_{\mathrm{rule}}=0.1$ following SDAR~\citep{lu2026self}. All checks use generated text and recorded histories without additional environment interaction.

\section{Additional Experiments}
\label{app:additional-experiments}

\subsection{Learning Dynamics}
\begin{figure*}[t]
    \centering
    \includegraphics[width=0.8\textwidth]{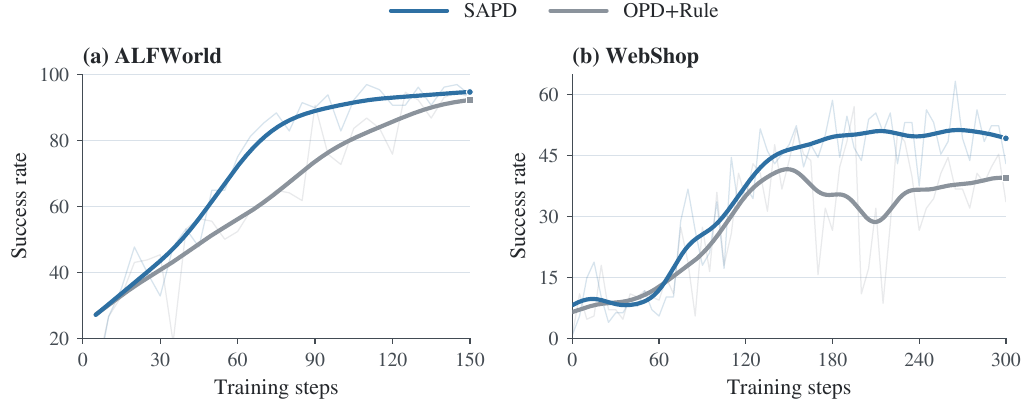}
    \caption{Evaluation success rates of SAPD and OPD+Rule
    during training with SFT teachers.}
    \label{fig:learning-dynamics}
\end{figure*}

To understand how SAPD's benefits develop over training, we compare its learning dynamics with OPD+Rule on ALFWorld and WebShop. Both methods train a Qwen2.5-1.5B-Instruct student using offline SFT prefixes and a frozen, SFT-trained Qwen2.5-3B-Instruct teacher. Figure~\ref{fig:learning-dynamics} shows checkpoint evaluation results together with smoothed trends. On ALFWorld, SAPD improves more rapidly during training, while SAPD maintains a higher success-rate trend during later training on WebShop. These observations suggest that SAPD supports faster learning progress on ALFWorld and stronger sustained performance on WebShop, complementing the results in Table~\ref{tab:main-results}.

\subsection{Distillation with Teacher-Generated Trajectories}
\label{app:grpo-trajectories}

Our main experiments focus on reusing existing offline trajectories, avoiding the need to collect teacher-specific rollouts through environment interaction. 
As a complementary analysis, we follow ReOPD~\citep{liao2026reopd} in reusing teacher-generated trajectories to construct decision prefixes, which are expected to better align with the teacher's interaction distribution and reduce teacher--prefix mismatch.
Although obtaining these trajectories requires environment interaction, subsequent student training remains environment-free. We compare SAPD and ReOPD on ALFWorld and WebShop using a 1.5B student and a GRPO-trained 3B teacher. As shown in Table~\ref{tab:grpo-trajectories}, SAPD achieves 76.6\% on ALFWorld and 70.8\% on WebShop, compared with 75.0\% and 66.7\% for ReOPD, yielding gains of 1.6 and 4.1 percentage points, respectively. These gains are smaller than those on SFT prefixes under the GRPO teacher in Table~\ref{tab:main-results}. These results suggest that SAPD remains effective on teacher-generated prefixes and may offer larger gains under stronger teacher--prefix mismatch.

\begin{table}[htbp]
\centering
\begingroup
\small
\setlength{\tabcolsep}{5pt}
\renewcommand{\arraystretch}{1.10}

\begin{tabular}{@{}l*{7}{r}c@{}}
\toprule
& \multicolumn{7}{c}{\textbf{ALFWorld}}
& \textbf{WebShop} \\
\cmidrule(lr){2-8}\cmidrule(l){9-9}
\textbf{Method}
& Pick & Look & Clean & Heat & Cool & Pick2 & \textbf{Avg.}
& \textbf{Success} \\
\midrule

ReOPD
& 92.2 & \textbf{61.1} & \textbf{93.0}
& 88.0 & 72.9 & \textbf{25.4} & 75.0
& 66.7 \\

\textbf{SAPD}
& \textbf{94.2} & 60.0 & 85.5
& \textbf{94.0} & \textbf{74.7} & 20.4 & \textbf{76.6}
& \textbf{70.8} \\

\bottomrule
\end{tabular}
\endgroup

\caption{Performance between SAPD and ReOPD. Avg. reports overall success across all evaluation episodes, giving greater weight to categories with more episodes. Best results are bold.}
\label{tab:grpo-trajectories}
\end{table}

\subsection{Distillation with a 3B Student}
\label{app:3b-context-mismatch}

\begin{figure}[t]
\centering
\includegraphics[width=0.56\linewidth]{
    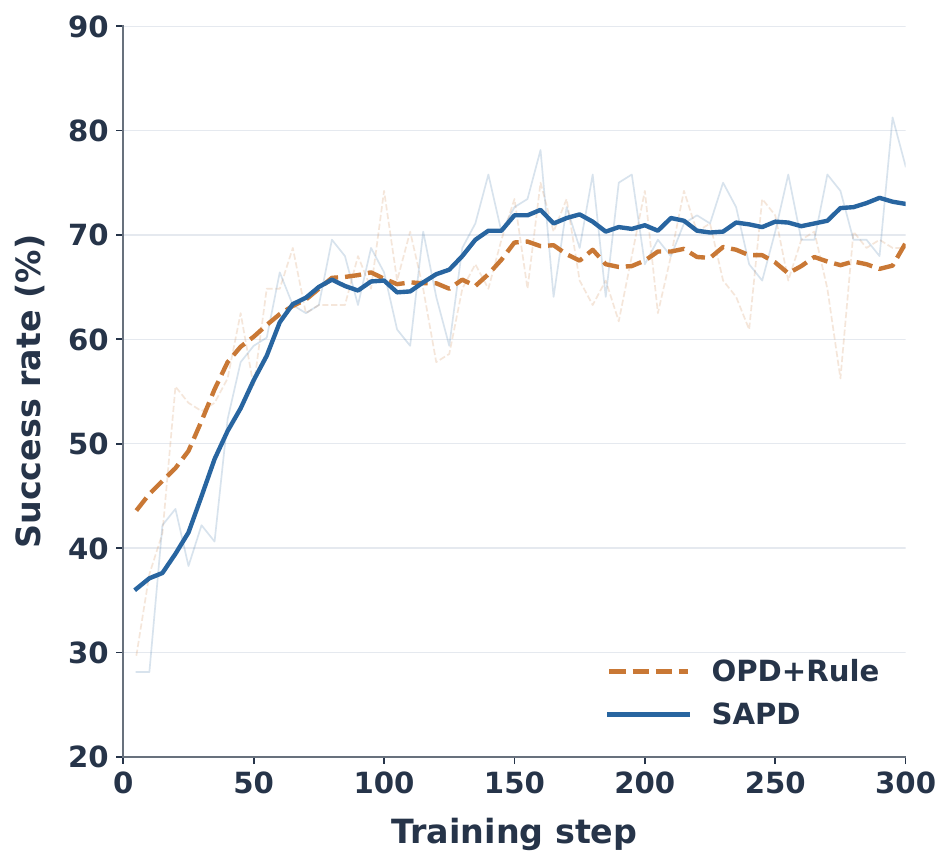
}
\caption{ALFWorld evaluation performance with a 3B student and a GRPO-trained 3B teacher using offline SFT prefixes.}
\label{fig:alfworld-3b-context-mismatch}
\end{figure}

Our main experiments focus on 1.5B students; we further investigate whether step-level guidance remains beneficial for a larger student under teacher--prefix mismatch. We evaluate a Qwen2.5-3B-Instruct student on ALFWorld using a GRPO-trained Qwen2.5-3B-Instruct teacher and decision prefixes extracted from offline SFT trajectories. Figure~\ref{fig:alfworld-3b-context-mismatch} compares OPD+Rule with SAPD, which computes relative advantages among candidate actions sharing the same prefix. Within 300 training steps, SAPD achieves the performance of 78.6\%, compared with 74.5\% for OPD+Rule. These results show that SAPD improves the performance of a 3B student under prefix mismatch, supporting its effectiveness in this setting.

\begin{table}[htbp]
\centering
\begingroup
\small
\setlength{\tabcolsep}{5pt}
\renewcommand{\arraystretch}{1.10}

\begin{tabular}{@{}l*{6}{c}@{}}
\toprule
& \multicolumn{3}{c}{\textbf{SFT teacher}}
& \multicolumn{3}{c}{\textbf{GRPO teacher}} \\
\cmidrule(lr){2-4}\cmidrule(lr){5-7}
\textbf{Scoring scope}
& ALFWorld & WebShop & Search
& ALFWorld & WebShop & Search \\
\midrule

Action-only
& \textbf{96.6} & \textbf{59.4} & \textbf{38.8}
& \textbf{69.3} & \textbf{53.4} & 41.3 \\

Full-token
& 95.9 & 50.0 & 38.6
& 60.8 & 52.3 & \textbf{41.6} \\

\bottomrule
\end{tabular}
\endgroup

\caption{Comparison of action-only and full-token scoring for step advantages. Both variants use offline SFT prefixes and update the complete response. Results are reported in percentages, with the better result in each setting shown in bold.}
\label{tab:scoring-scope}
\end{table}

\section{Ablation on Step-Advantage Scoring}
\label{app:scoring-scope}
A generated response contains both reasoning text and an executable action, raising the question of whether step advantages should be computed from action tokens alone or from all response tokens. We compare these two scoring scopes using offline SFT trajectories under both SFT and GRPO teachers. Both variants apply policy updates to the complete response; only the tokens used to compute step scores differ. As shown in Table~\ref{tab:scoring-scope}, action-only scoring performs better on ALFWorld and WebShop under both teacher configurations. It improves ALFWorld success from 60.8\% to 69.3\% with the GRPO teacher and WebShop success from 50.0\% to 59.4\% with the SFT teacher. On Search-based QA, the differences are small. These results support action-focused scoring and suggest that excluding thinking tokens may provide more reliable step-level guidance when prefix mismatch makes teacher supervision less reliable.

\end{document}